\documentclass[runningheads]{llncs}

\usepackage{framed,multirow}
\usepackage{rotating}
\usepackage{lscape}
\usepackage{booktabs}
\usepackage{amssymb}
\usepackage{amsmath}
\usepackage{latexsym}
\usepackage{float}
\usepackage{algorithm}
\usepackage{lineno}
 \usepackage{algpseudocode}
\usepackage{setspace}
\usepackage{url}
\usepackage{xcolor}
\usepackage{caption}
\usepackage{subcaption}
\usepackage[margin=2cm]{geometry}
\usepackage{hyperref}

\usepackage[T1]{fontenc}

\usepackage{natbib}
\setcitestyle{numbers}

\begin{document}

\title{\textit{Varroa destructor} detection on honey bees using hyperspectral imagery}
%
%
\author{Zina-Sabrina Duma\inst{1} \and
Tomas Zemcik\inst{2} \and Simon Bilik\inst{1,2} \and
Tuomas Sihvonen\inst{1} \and
Peter Honec\inst{2} \and Satu-Pia Reinikainen\inst{1} \and Karel Horak\inst{2}}
\authorrunning{Duma et al.}

%
\institute{LUT University, Yliopistonkatu 34, Lappeenranta 53850 Finland, \textit{Zina-Sabrina.Duma@lut.fi} \and
Brno University of Technology, Faculty of Electrical Engineering and Communication, Technická 3058/10, Brno 61600, Czech Republic}

    

%
\maketitle              
\begin{abstract}
Hyperspectral (HS) imagery in agriculture is becoming increasingly common. These images have the advantage of higher spectral resolution. Advanced spectral processing techniques are required to unlock the information potential in these HS images. The present paper introduces a method rooted in multivariate statistics designed to detect parasitic \textit{Varroa destructor} mites on the body of western honey bee \textit{Apis mellifera}, enabling easier and continuous monitoring of the bee hives. The methodology explores unsupervised (K-means++) and recently developed supervised (Kernel Flows - Partial Least-Squares, KF-PLS) methods for parasitic identification. Additionally, in light of the emergence of custom-band multispectral cameras, the present research outlines a strategy for identifying the specific wavelengths necessary for effective bee-mite separation, suitable for implementation in a custom-band camera. Illustrated with a real-case dataset, our findings demonstrate that as few as four spectral bands are sufficient for accurate parasite identification.

\keywords{hyperspectral imagery (HSI) \and varroa destructor \and beehive monitoring \and wavelength selection \and kernel partial least-squares}
\end{abstract}

\section{Introduction}

Environmental issues, alongside the use of pesticides and the presence of parasites, pose significant threats to bee populations worldwide. Among these threats, the \textit{Varroa Destructor} (Varroa) mite is particularly notorious for its role in most instances of Colony Collapse Disorder (CCD), as highlighted in recent studies \cite{flores2021impact, eliash2020varroa}.

Traditionally, detecting Varroa mites within beehives has relied on manual, non-automated methods such as sugar shake tests, brood examinations, and debris analysis \cite{jack2021integrated, roth2020biology}. The recent advancements have introduced computer vision techniques (CV) utilizing the automated analysis of the bee debris on a monitoring plate in~\cite{konig2020varroacounter}, or directly on the bee's body using conventional CV techniques in~\cite{BJERGE2019104898}, convolutional neural network (CNN) classifiers in~\cite{9897809} or the deep object detectors in~\cite{bilik2021visual, liu2023detection}. An extensive overview of the computer vision techniques used for the Varroa mite monitoring and bee colony health monitoring, in general, is shown, for example, in~\cite{BILIK2024108560}. Besides the CV techniques, analysis based on the sensor and sound data is used, as presented, e.g. in~\cite{hall2023automated,mekha2022honey}. Nevertheless, reliable detection of the Varroa mite in the visible spectrum is challenging as it appears similar to the bee's body or the surroundings.


Hyperspectral (HS) imagery for agriculture monitoring is becoming readily available \cite{lu2020recent}. There is an increased demand for precise performant HS imaging processing techniques \cite{khan2022systematic}. For monitoring bees and insects, HS imagery has been utilized previously in very few cases. More studies are found with multispectral data. The authors of~\cite{BJERGE2019104898} used the VideometerLab4 instrument sensing in 19 wavelengths in the range of 375-970~nm to measure samples of the bees and Varroa mites followed by Linear Discriminant Analysis (LDA) on the spectral data to design an optimal illumination for their bee monitoring device; and the authors of~\cite{maanefjord20223d} used the multispectral data for insect monitoring, but not for Varroa mite detection.

In this study, we introduce a novel approach for extracting spectral data from HS images and leveraging this data to calibrate spectral signatures for identifying clusters of parasites. We also propose methods for selecting wavelengths that discriminate between clusters. The objective of this research is to explore the potential of HS imagery in addressing significant questions related to bee health and parasite detection:

\begin{itemize}
    \item \textit{Is it possible to use hyperspectral imagery to identify Varroa mites on bees?}
    \item \textit{What procedures are necessary to extract discriminative information between bees and Varroa mites?}
    \item \textit{How many and which specific wavelengths are crucial for distinguishing between bees and Varroa mites?}
    \item \textit{Can statistical-based methods yield reliable results with independent data sets?}
\end{itemize}

Our methodology encompasses a process of spectral reconstruction aimed at enhancing the differentiation between bees and Varroa mites, which is crucial for supporting classification or clustering algorithms. This approach accentuates the contrasts between bee-mite characteristics while minimizing the variations caused by background elements, shadows, and pixel noise. Pixel noise, in particular, is often a byproduct of the line-scanning technique employed by hyperspectral cameras \cite{bjorgan2015real}. The process involves utilizing Principal Component Analysis (PCA) \cite{rodarmel2002principal} to decompose the HS image, followed by selecting only those principal components that demonstrate a strong absolute correlation with a bee-mite discrimination variable for the image reconstruction. K-means++ method \cite{hamalainen2020improving} is applied to cluster the reconstructed HS images.  Cluster centers are being applied to new images for effective discrimination.

Kernel Flows - Partial Least-Squares (KF-PLS)\cite{duma2023kf} was utilized in the case of supervised clustering. It has the property to maintain the qualities of multivariate statistical methods, such as the reduced size of the calibration dataset, while extending applicability to non-linear relationships. The spectra of bees and parasites are not expected to be distinguished in a fully linear method; thus, there is a need for spectral projection in a Reproducing Kernel Hilbert Space, where they become linearly separable. 

For the selection of spectral bands essential to discrimination between bees and Varroa mites, the mathematical methods utilized are two partial least-squares (PLS) \cite{de1993simpls} based methods: a modified version of the Covariance Procedure \cite{reinikainen2003covproc}, mentioned throughout the paper as the COVPROC method, and the explained variance by wavelength (referred to as the $R^2$ method), derived from forward interval partial least-squares (FiPLS) \cite{yun2019overview}.

The methodologies outlined in this paper were evaluated using a two-part proprietary hyperspectral image dataset. This dataset is made accessible alongside this article.


The innovation of this research lies in several key contributions: (\textit{i)} introducing the use of HS imagery for detecting Varroa mites, a novel application in the field; (\textit{ii}) developing a procedure for reconstructing spectral data that effectively excludes background details, shadows, and noise, ensuring clearer differentiation between subjects of interest; (\textit{iii}) offering a new methodology for identifying discriminating spectral bands, which enhances understanding of the specific wavelengths vital for distinguishing between bees and Varroa mites; (\textit{iv}) proposing a technique for creating spectral profile-based classifiers, which could improve how spectral data is used for classification purposes; (\textit{v)} demonstrating the feasibility of using minimal data for training, addressing one of the common challenges in machine learning applications by reducing the requirement for extensive training datasets.

\section{Materials and Methods}

This section presents the origin of bees and Varroa mites samples together with their measurement arrangement (Subsection~\ref{ssec:insects}), followed by the instrumentation, measurement setup and image data (Subsection~\ref{ssec:dataInstrumentation}), and summarized with the mathematical procedures and algorithms used for processing of the collected data (Subsection~\ref{ssec:mathematicalMethods}).

\subsection{Collected samples of bees and Varroa mites}\label{ssec:insects}

Two batches of samples were collected from different areas and at different times for the study. The first samples were taken from Těšínky (49.9566236N, 15.1436481E; CZ) in November 2021. The second was collected from Kroměříž (49.3005392N, 17.3797958E; CZ) in June 2022. These locations are situated approximately 180~km from one another. In both cases, the detritus containing dead mites and bees was collected from the bottom of the hives. In addition, detritus from the Těšínky locality was collected after a regular autumn fumigation with Amitrazum 125 mg/ml.

\begin{figure}[H]
     \centering
     \begin{subfigure}[t]{0.3\textwidth}
         \centering
         \includegraphics[width=\textwidth]{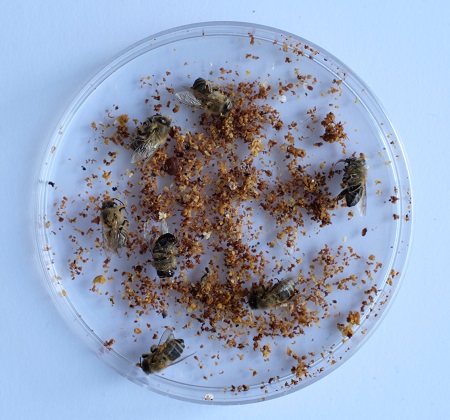}
         \caption{Bees mixed-up with detritus.}
         \label{fig:DS11}
     \end{subfigure}
     \begin{subfigure}[t]{0.3\textwidth}
         \centering
         \includegraphics[width=\textwidth]{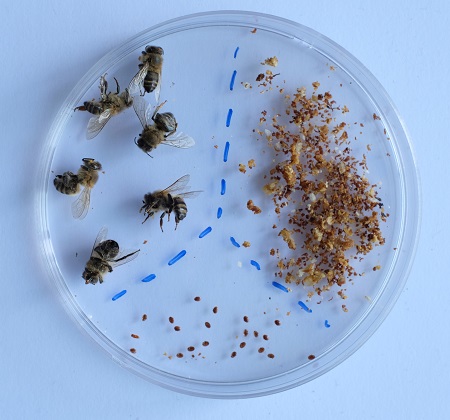}
         \caption{Clockwise: Bees; detritus; Varroa mites.}
         \label{fig:DS12}
     \end{subfigure} 
    \begin{subfigure}[t]{0.3\textwidth}
         \centering
         \includegraphics[width=\textwidth]{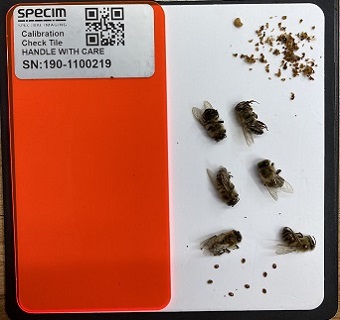}
         \caption{Sample with detritus, bees, bees with Varroa mites and Varroa mites on the calibration plate.}
         \label{fig:DS21}
     \end{subfigure}
     \begin{subfigure}[t]{0.3\textwidth}
         \centering
         \includegraphics[width=\textwidth]{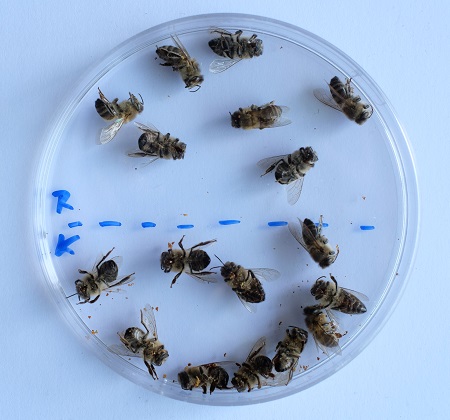}
         \caption{Upper part: freshly collected bees; Lower part: 3 months old bee samples.}
         \label{fig:DS13}
     \end{subfigure}
     \begin{subfigure}[t]{0.3\textwidth}
         \centering
         \includegraphics[width=\textwidth]{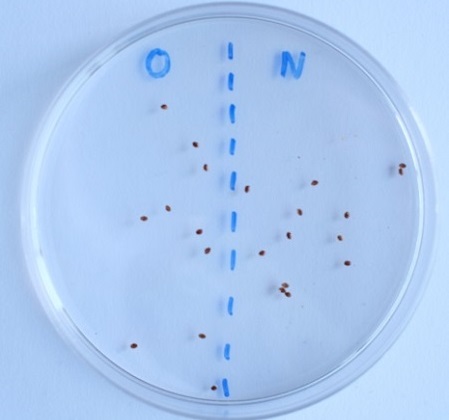}
         \caption{Left part: 3 month old Varroa mites; Right part: freshly collected Varroa mites.}
         \label{fig:DS14}
         \end{subfigure}
     \begin{subfigure}[t]{0.3\textwidth}
         \centering
         \includegraphics[width=\textwidth]{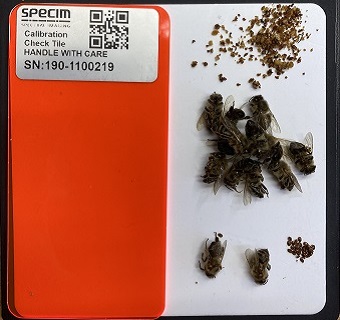}
         \caption{Sample with clustered detritus, bees, bees with Varroa mites and Varroa mites on the calibration plate.}
         \label{fig:DS22}
         
     \end{subfigure}
    \caption{Samples from the HS dataset utilizied in calibration (a,b, d, e) and testing (c, f).}
    \label{fig:DS}
\end{figure}

The first set of samples, illustrated in Fig.~\ref{fig:DS11}, \ref{fig:DS12}, \ref{fig:DS13} and \ref{fig:DS14} contains samples arranged on the Petri dishes organized as follows:

\begin{itemize}
    \item Bees mixed up with detritus from locality Kroměříž (Fig.~\ref{fig:DS11}).
    \item Separated bees, detritus and Varroa mites from locality Kroměříž (Fig.~\ref{fig:DS12}).
    \item Fresh bees from locality Kroměříž, marked as \textit{K}, and eight months old bees from locality Těšínky, marked as \textit{R} (Fig.~\ref{fig:DS13}).
    \item Eight months old Varroa mites from locality Těšínky, marked as \textit{O} and fresh Varroa mites from locality Kroměříž marked as \textit{K} (Fig.~\ref{fig:DS14}).
\end{itemize}

To suppress the effect of the polystyrene Petri dish used in HS images of the first set, a second set was prepared out of the Kroměříž samples placed directly on a white reference panel supplied with the used camera, which should have constant spectral properties across the camera's spectral range. Images of the detritus, bees, Varroa mites and bees with placed Varroa mites were taken in a dense and clustered arrangement as shown in right column of Fig.~\ref{fig:DS}, with the dense arrangement shown in Fig.~\ref{fig:DS21} and clustered arrangement shown in Fig.~\ref{fig:DS22}.

In addition to the mites and the bees, the detritus consists of any kind of waste material from the nest, such as wax, pollen, or sugar, which is also visible in Fig.~\ref{fig:DS11}, \ref{fig:DS12}, \ref{fig:DS21} and~\ref{fig:DS22}. We also made our dataset publicly available in~\cite{BeeDS_HS}.


\subsection{Hyperspectral imagery}\label{ssec:dataInstrumentation}

The hyperspectral images were taken on a Specim IQ ~\cite{Behmann2018specimIQ} portable hyperspectral camera that allows a simple measurement setup and fast acquisition. Camera parameters relevant to the data format are included in Table~\ref{tab:DataParameters}.

\begin{figure}[H]
     \centering
     \begin{subfigure}[t]{0.4\textwidth}
         \centering
         \includegraphics[width=\textwidth]{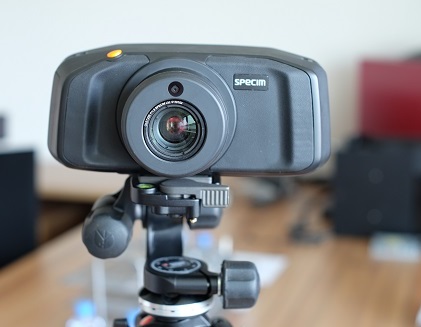}
         \caption{Used Specim IQ camera.}
         \label{fig:Spec}
     \end{subfigure}
     \begin{subfigure}[t]{0.4\textwidth}
         \centering
         \includegraphics[width=\textwidth]{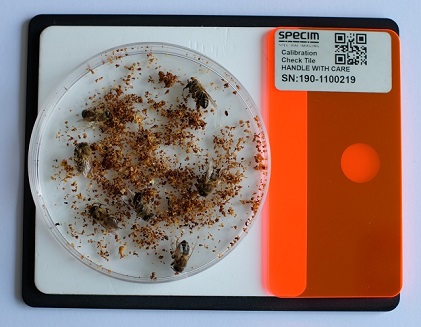}
         \caption{Sample on the calibration plate.}
         \label{fig:MSetup}
     \end{subfigure}
        \caption{Utilised camera and measurement setup.}
        \label{fig:Setup}
\end{figure}

\begin{table}[ht!]
	\begin{center}
        \begin{tabular*}{0.5\textwidth}{@{\extracolsep{\fill}}ll}
        \toprule
        \multicolumn{2}{c}{Specim IQ image parameters} \\
        \midrule 
        Spatial resolution       & 512 px x 512 px     \\
        Field of view            & 31° x 31°           \\
        F/number                 & 1/1.7               \\
        Min. focus distance      & 150 mm              \\
        Scanning principle       & push-broom          \\
        Spectral range           & 400 nm - 1000 nm    \\
        Spectral resolution      & 7 nm                \\
        Spectral bands           & 204                 \\
        Image format             & ENVI compatible     \\
        \bottomrule
    \end{tabular*}
    \end{center}
\caption{Basic camera and image parameters of Specim IQ ~\cite{Behmann2018specimIQ, Ikäheimo2018SpecimIQ, Zemcik2023Inks}}
\label{tab:DataParameters}
\end{table}

The samples were illuminated with a multispectral unit consisting of 29 LEDs with individually controllable drivers. The LEDs for the unit were selected to cover the spectral range of the experiments. Dedicated software can adjust the illumination to optimize the lighting of the samples. For these experiments, the illumination was set up for approximately the inverse spectrum relative to the sensitivity of the Specim IQ camera. Each image in the dataset includes a spectral calibration target with known spectral properties that allow for rectification of the spectra as shown in Fig.~\ref{fig:MSetup}.


For the modelling goal, a group of images from the dataset was selected, containing two sets of images of dead bees and Varroa mites collected from bee detritus, a total of six images. The first image set was used to calibrate the models, whereas the second was held out for testing purposes.

\begin{figure}[H]
    \centering
    \begin{subfigure}[b]{0.3\linewidth}
        \centering
        \includegraphics[width=\linewidth]{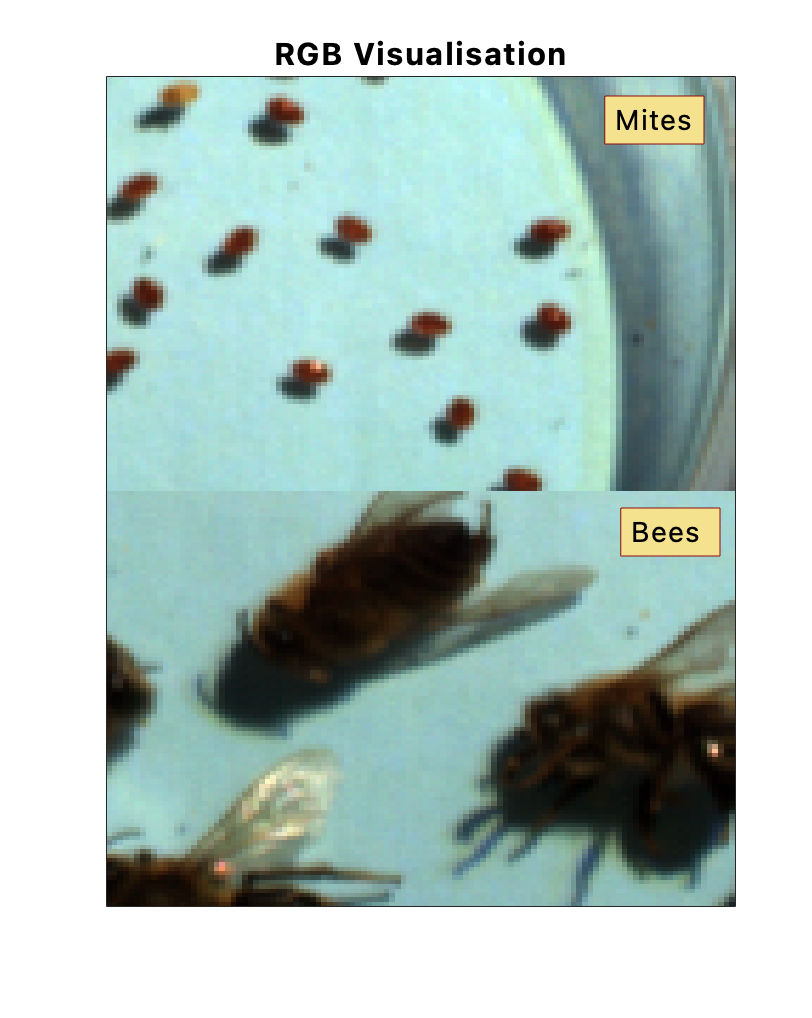}
        \caption{}
        \label{fig:calibrationImage}
    \end{subfigure}
    \begin{subfigure}[b]{0.49\linewidth}
        \centering
        \includegraphics[width=\linewidth]{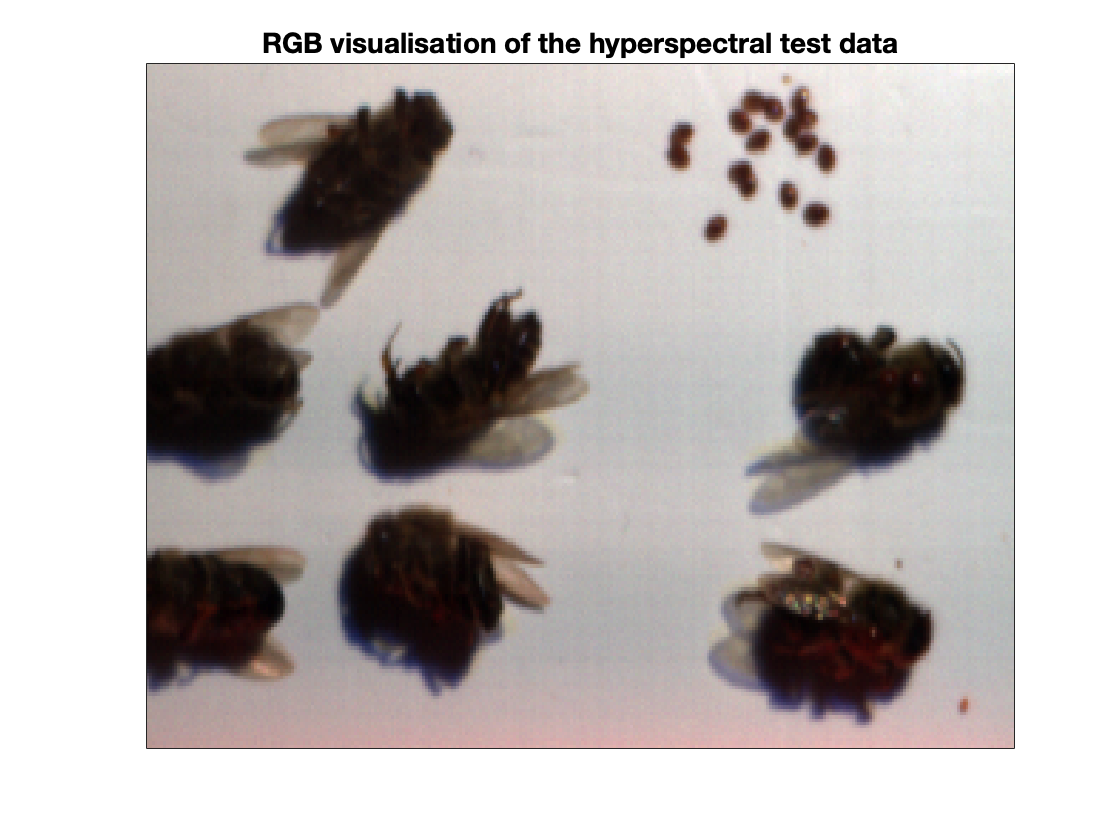}
        \caption{}
        \label{fig:testImage}
    \end{subfigure}
    \caption{RGB visualisations of the hyperspectral images with 204 bands for the (a) model calibration data and (b) model testing data.}
    \label{fig:dataVisualisation}
\end{figure}

Fig.~\ref{fig:dataVisualisation} showcases RGB visualisations of the 204-wavelength images utilized for developing the model. The calibration set in Fig.~\ref{fig:calibrationImage} has separate bees and Varroa mites placed on a petri dish on top of a white paper sheet. The testing hyperspectral image is present in Fig.~\ref{fig:testImage} and presents bees (left), Varroa mites (up, right), and bees that have Varroa mites on top of them (down, right). The goal of the model is not only to discriminate between the bees and Varroa mites but also to correctly identify the Varroa mites on top of a bee - a more likely scenario in real-case studies. Another notable difference between the calibration and testing sets is the different backgrounds, which can also differ in real-case scenarios. 

\subsection{Mathematical methods and workflow}\label{ssec:mathematicalMethods}

This subsection describes the spectral reconstruction and clustering methods, followed by the wavelength selection technique and the algorithms used.

\subsubsection{Spectral reconstruction and clustering}

\begin{figure}[H]
    \centering
    \includegraphics[width=0.4\linewidth]{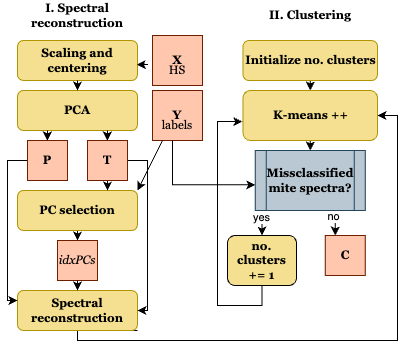}
    \caption{Cluster formation workflow.}
    \label{fig:workflow}
\end{figure}

The present section presents the mathematical methods building blocks for the workflow in Fig~\ref{fig:workflow}. 
The first procedure is to center and scale the spectra by subtracting the wavelength mean and dividing by its standard deviation, as seen in Eq~\ref{eq:centerscale}, where $\textbf{x}_i$ is an individual wavelength \textit{i}, $\hat{x}_i$ is the mean of the wavelength and $\sigma_i$ its standard deviation. Centering ensures that the largest variation profile does not mimic the spectral average and that principal components (PCs) pass through the origin, and scaling ensures that wavelengths are given the same importance in the model. 

\begin{equation}
    \textbf{x}_i = \frac{\textbf{x}_{raw, i} - \hat{x}_i}{\sigma_i}
    \label{eq:centerscale}
\end{equation}

In PCA, the scaled and centred spectra are decomposed into spectral profiles named Principal Components (PCs). The method can be summarized by Eq.~\ref{eq:PCA}, where \textbf{T} is the score matrix, \textbf{P} is the loadings matrix and \textbf{E} is the residual matrix.

\begin{equation}
    \textbf{X} = \textbf{T} \textbf{P}^T + \textbf{E}
    \label{eq:PCA}
\end{equation}

The reconstruction of spectra is made only with the PCs whose scores $\textbf{t}$ have a high absolute correlation with the discriminating variable \textbf{y}. This ensures that variational profiles that discriminate between background and insects or principal components related to noise are not included. The pixels that are Varroa mites or bees are extracted from the calibration data based on the calibration set ground truth. These pixels' scores are then evaluated in relation to a dummy variable with two possible values: '1' in case of membership to the 'bee' class and '0' in case of membership to the 'mite' class. The absolute correlation coefficients are calculated as in Eq.~\ref{eq:corrCoeff}, where $\textbf{t}_i$ is the score vector of selected pixels for the \textit{i}-th PC.

\begin{equation}
    \rho(\textbf{t}_i, \textbf{y}) = \frac{cov(\textbf{t}_i, \textbf{y})}{\sigma_{t_i}\sigma_{\textbf{y}}}
    \label{eq:corrCoeff}
\end{equation}

The reconstructed spectra are then subjected to a modified K-means++ algorithm with the methodology presented in Duma \textit{et al.}, 2023 \cite{duma2023colorimetric}. The initial clustering is considered correct if there are no 'mite' pixels miss-classified, meaning there are no false alarms, pixels inside bees classified as Varroa mites, or un-detected Varroa mites. If one or the previous cases occurs, the number of clusters is increased until the criteria are met.

To utilize the calibrated clusters \textbf{C} with a new image, the spectra need to be scaled, centered using the calibration mean and center, and then projected into the PCA model, as in the following set of equations:

\begin{equation}
\begin{split}
    \textbf{X}_{new} &= \frac{\textbf{X}_{raw, new} - \hat{\textbf{x}}}{\boldsymbol{\sigma}} \\
    \textbf{T}_{new} &= \textbf{X}_{new}  \textbf{P} \\
    \hat{\textbf{X}}_{new} &= \textbf{T}_{new, sel} \textbf{P}_{sel}^T
\end{split}
\end{equation}

The Euclidean distance of the newly reconstructed spectra $\hat{\textbf{X}}_{new}$ is then evaluated to the calibration centroids \textbf{C}, and pixel membership to a cluster is assigned.

In the case of unsupervised clustering, K-means++ was chosen for its ease of use. While K-means is a performant method for unsupervised clustering, it needs to be provided with spectral profiles for efficient cluster identification, as per the methodology presented below; otherwise, it is affected by background variation, shadows, or noise.

\begin{figure}[H]
    \centering
    \includegraphics[width=0.5\linewidth]{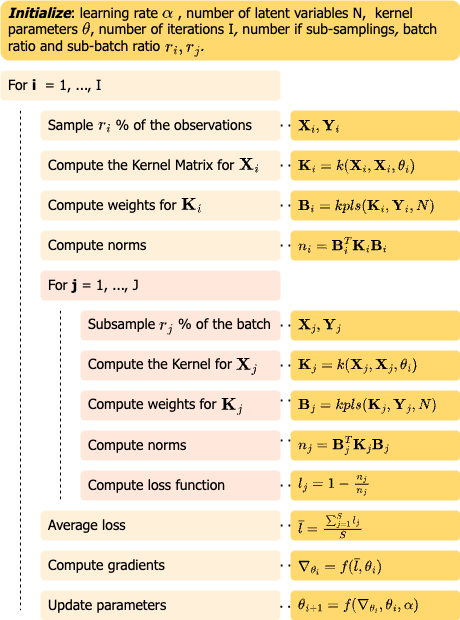}
    \caption{KF-PLS workflow.}
    \label{fig:KFPLSWorkflow}
\end{figure}

However, if one has access to at least a set of labelled data and would like to input the spectra as it is, without spectral profiling, an alternative would be the usage of a multivariate statistical method with discriminant analysis, such as the Kernel Flows - Partial Least-Square (KF-PLS) \cite{duma2023kf} with Discriminant Analysis. KF-PLS is a variant of optimized Kernel PLS \cite{rosipal2001kernel}, where the wavelengths are projected via a Kernel function (Gaussian, Laplacian, Matern or Cauchy) into a Reproducing Kernel Hilbert Space via a Kernel Trick \cite{wu2005formulating}. If the relationship between the spectra and the response variable (classes) is non-linear in the original space, a higher dimension is found where the relationship spectra-classes these becomes linear \cite{cook2021pls}. The optimization is based on Kernel Flows \cite{owhadi2019kernel} that learns the kernel parameters in a cross-validation manner. The Kernel Flows version of K-PLS is an appropriate choice for the bee-mite discrimination case because (a) it is capable of self-learning the kernel parameters to suit the case, and (b) it can utilize as little information as possible, that includes the necessary information for discrimination, being based in multivariate statistical methods. The workflow of KF-PLS is presented in Fig.~\ref{fig:KFPLSWorkflow}. 

\subsubsection{Wavelength selection}

Two types of methods based on Partial Least-Squares (PLS) regression have been considered for the wavelength selection. In this study, the SIMPLS version of PLS was utilized \cite{de1993simpls}. In PLS, the response matrix \textbf{Y} is also decomposed, as in Eq.~\ref{eq:PLS}, where \textbf{U} is the \textbf{Y}-side score matrix, \textbf{Q} is the \textbf{Y}-side loadings matrix and \textbf{F} is the \textbf{Y}-side residual matrix. In this scenario, the PLS version utilized is PLS with Discriminant Analysis (PLS-DA), as the \textbf{y} variables give information on the membership or non-membership of a pixel to a cluster.

\begin{equation}
    \textbf{Y} = \textbf{U} \textbf{Q}^T + \textbf{F}
    \label{eq:PLS}
\end{equation}

The first wavelength selection method is the iterative forward selection based on the PLS model between the wavelengths and the group membership variable \textbf{y}. The variable selection method is a modified version of the Forward Interval PLS (FiPLS) \cite{balabin2011variable}, where instead of utilizing intervals and cross-validation, the search was exhaustive throughout all variables. To initialize the selection of variables, the first three correlated variables can be calculated with the correlation coefficient with respect to the response variable, as in Eq.~\ref{eq:corrCoeff}.

The regression coefficients can be obtained through Eq.~\ref{eq:bPLS}, where \textbf{W} is the matrix of rotated loadings in the direction of maximum \textbf{X} and \textbf{y} covariance.

\begin{equation}
    \mathbf{b} = \mathbf{W} (\mathbf{P}^T \mathbf{W})^{-1} \mathbf{Q}^T
    \label{eq:bPLS}
\end{equation}

After the initial selection, the following steps can be followed, as seen in \textbf{Algorithm} \ref{alg:R2varSelect}: (\textit{i}) add the variables one at a time to the selected list of variables, (\textit{ii}) perform PLS between the selected variables and the response, (\textit{iii}) evaluate which iteration had the highest explained variance of the \textbf{y}-variable $R^2$, (\textit{iv}) permanently add to the selected list of variables the wavelength whose addition yielded the highest $R^2$, (\textit{v}) repeat Step \textit{i} until the desired number of variables has been collected. In the present case, adding variables stops when the success conditions of the clustering algorithm are met: no 'mite' pixels are misclassified. 

\begin{algorithm}[H]
\caption{$R^2$-based Variable Selection}\label{alg:R2varSelect}
\textbf{\textit{Input}}: spectral matrix, with wavelengths as columns and bee-mite pixels as rows ($\mathbf{X}$), bee-mite discriminating vector ($\mathbf{y}$), the desired number of variables to be selected ($V$). \newline \textbf{\textit{Output}}: vector with selected wavelength indices of (\textbf{s}). \newline
\textbf{\textit{Initialize}}: initial vector of selected variables ($\mathbf{s}_0$), initial vector of unselected variables ($ \mathbf{sn}_0 $), initial number of selected ($N$) and unselected ($M$) wavelengths.
\begin{algorithmic}[1]
\For{$v \gets 1$ to $V$} \Comment{Loop until desired number of variables selected.}
\State $   TSS \gets \sum (\mathbf{y} - \bar{\mathbf{y}})^2 $ \Comment{Calculate the total sum of squares.}  
\For{$m \gets 1$ to $M$} \Comment{Loop through unselected variables.}
\State $ \mathbf{s}_{temp} \gets [ \mathbf{s}, \mathbf{sn}_m]$ \Comment{Add the current unselected variable to the temporary variable list.}
\State $ \mathbf{b} \gets pls(\mathbf{X}_{s_{temp}}, \mathbf{y}) $ \Comment{Calibrate PLS model.}
\State $ \hat{\mathbf{y}} \gets \mathbf{X}_{s_{temp}} \mathbf{b}$ \Comment{Estimate \textbf{y} from the model.}
\State $ RSS_m \gets \sum (\mathbf{y} - \hat{\mathbf{y}})^2$ \Comment{Calculate the prediction sum of squares.}
\State $ R^2_m \gets 1 - \frac{RSS_m}{TSS}$ \Comment{Calculate the prediction $R^2$ for the iteration.}
\EndFor
\State $ idx \gets max(\textbf{R}^2)$ \Comment{Select the iteration with the highest increase.}
\State $ sn_{N+1} \gets sn_{idx} $ \Comment{Add variable to selected variables.}
\State $ N \gets N + 1 $ \Comment{Increase the counter of selected variables.}
\EndFor
\end{algorithmic}
\end{algorithm}

The second version of wavelength variable selection proposed is a modified version of the Covariance Procedures (COVPROC) \cite{reinikainen2007multivariate}, that is presented in \textbf{Algorithm}~\ref{alg:COVPROC}. An optional step of relevance to the present paper is sorting the rounds based on their included number of variables. 

\begin{algorithm}[H]
\caption{COVPROC-based Variable Selection}\label{alg:COVPROC}
\textbf{\textit{Input}}: spectral matrix, with wavelengths as columns and bee-mite pixels as rows ($\mathbf{X}$), bee-mite discriminating vector ($\mathbf{y}$), the desired number of 
COVPROC rounds ($R$), the total number of wavelengths ($I$).
\newline
\textbf{\textit{Output}}: selected list of variables ($\textbf{s}$) \newline
\textbf{\textit{Initialize}}: a vector with '0' values of length I (\textbf{a})
\begin{algorithmic}[1]
\For{$r \gets 1$ to $R$}
    \State $ \mathbf{w} \gets pls(\mathbf{X}, \mathbf{y})$ \Comment{Extract PLS weights for one latent variable.}
    \State $ \mathbf{idxw} \gets sort(|\mathbf{w}|)$ \Comment{Sort the weights in descending order and save indexes.}
    \State $ \mathbf{w}_r \gets \mathbf{a}$ \Comment{Reset $\textbf{w}_b$}
    \For{$i \gets 1$ to $I$}
        \State $ ii \gets idxw_i $ \Comment{Select variable.}
        \State $ \textbf{w}_{r, ii} \gets \textbf{y}^T \textbf{x}_{ii}$ \Comment{Populate $\textbf{w}_r$ with selected variable/ response covariance.}
        \State $ \mathbf{t}_{r,i} \gets \mathbf{X} \mathbf{w}_r $ \Comment{Compute iteration's scores.}
        \State $ \alpha_i \gets \frac{|\mathbf{y}^T \mathbf{t}_{r,i} |}{\mathbf{t}_{r,i}^T \mathbf{t}_{r,i}} $ \Comment{Calculate evaluation metric.}
    \EndFor
    \State $ n \gets max(\boldsymbol{\alpha}) $ \Comment{Extract the index of the maximum $\boldsymbol{\alpha}.$}
    \State $ \mathbf{s}_r \gets \mathbf{idxw}_{1, ..., n}$ \Comment{Extract the variables of the round.}
    \State $ \mathbf{s} \gets [\mathbf{s}, \mathbf{s}_r] $ \Comment{Append the selected variable list of round \textit{r}.}
    \State $ \mathbf{p}_r \gets \frac{\mathbf{X}^T \mathbf{t}_{r, n}}{\mathbf{t}_{r, n}^T \mathbf{t}_{r, n}}$ \Comment{Compute the loadings of the round \textit{r}.}
    \State $ \textbf{X} \gets \mathbf{X} - \mathbf{t}_{r, n} \mathbf{p}_r^T $ \Comment{Deflate \textbf{X}.}
\EndFor

\end{algorithmic}    
\end{algorithm}

\section{Results and Discussion}
\subsection{Variation profiles}

\begin{figure}[H]
    \centering

    \begin{subfigure}[b]{0.49\linewidth}
        \centering
        \includegraphics[width=0.85\linewidth]{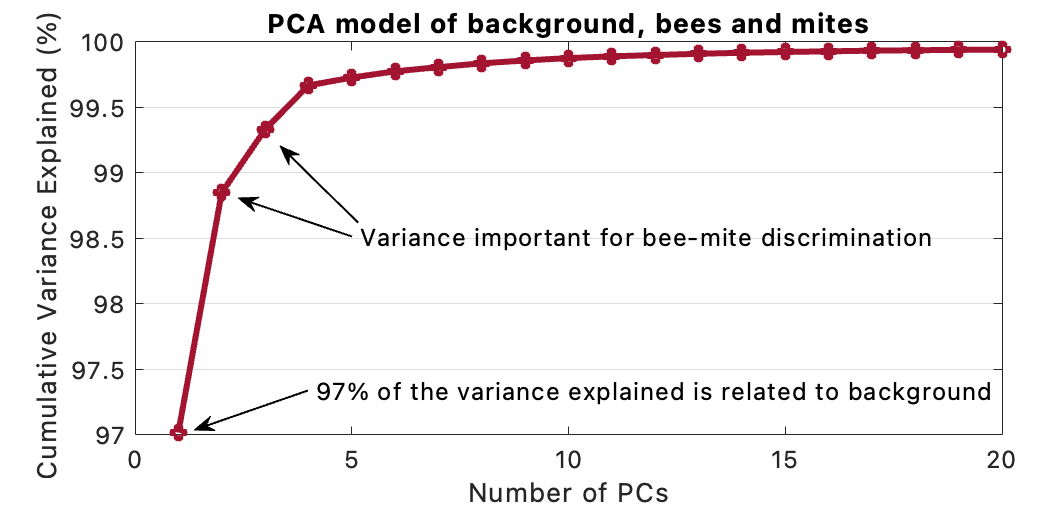}
        \caption{}
        \label{fig:PCvariationExplained}
    \end{subfigure}
    \hfill
    \begin{subfigure}[b]{0.49\linewidth}
        \centering
        \includegraphics[width =\linewidth]{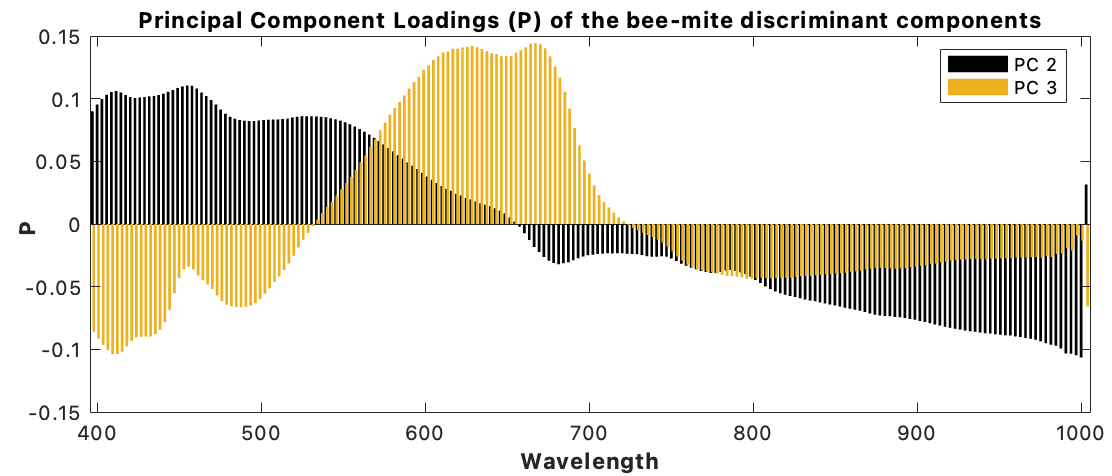}
        \caption{}
        \label{fig:PCLoadings}
    \end{subfigure}
    \begin{subfigure}[b]{0.8\linewidth}
        \includegraphics[width = \linewidth]{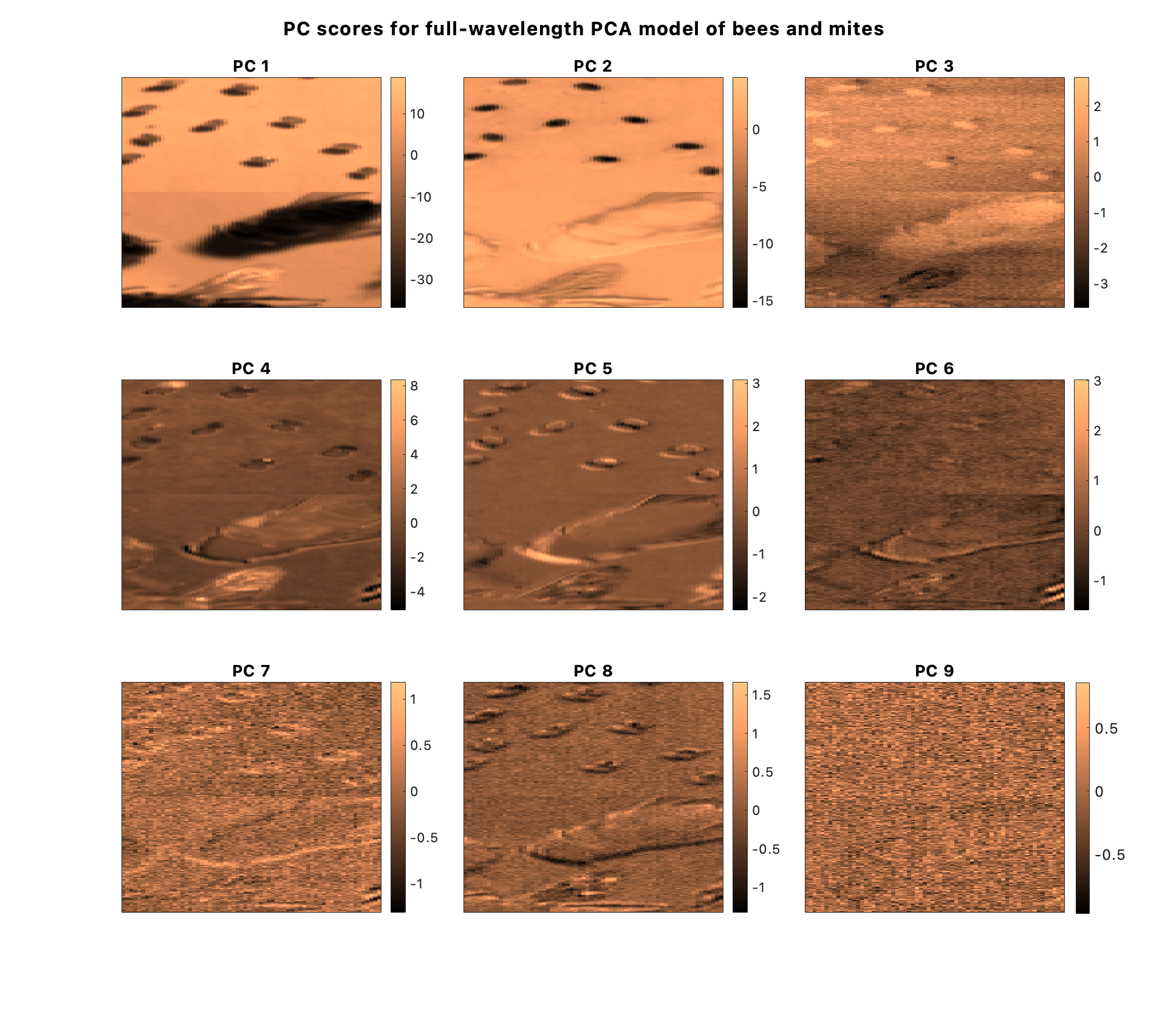}
        \caption{}
        \label{fig:PCScores}
    \end{subfigure}
    
    \caption{(a) Explained variance of calibration dataset for the principal components model. (b) Score values for the first 10 PCs. The PCs responsible for distinguishing between the bees and Varroa mites are PC1 and PC2, and the loadings of the discriminant PCs (c).}
    \label{fig:PCAmodel}
\end{figure}

After centring and scaling the image, the principal component analysis revealed that the background of the hyperspectral images represents 97\% of the systematic variation, as seen in Fig.~\ref{fig:PCvariationExplained} and Fig.~\ref{fig:PCScores}. One can observe that the 2nd and 3rd principal components, responsible for capturing the systematic variation between bees and Varroa mites, can be utilized for spectral reconstruction. The two PCs sum up to approximately 2\% of the total variation. These were also the principal components whose scores had the highest absolute correlation coefficient with the bee-mite discriminating variable and thus were selected for reconstruction. 

The loadings of the discriminating PCs are displayed in Fig.~\ref{fig:PCLoadings}. It is observable that the 2nd principal component has high loadings at the beginning and end of the spectrum, whereas the second one describes the middle part of the spectra. The higher the absolute value of a wavelength loading to a principal component, the more important it is for the variation profile. The rest of the principal components seem to either explain variation related to wings (PC4), shadows (PC4 to PC8), or noise (over PC9).  

\subsection{Full-spectral clustering}

\begin{figure}[H]
    \centering
    \begin{subfigure}[b]{0.15\linewidth}
        \centering
        \includegraphics[width =\linewidth]{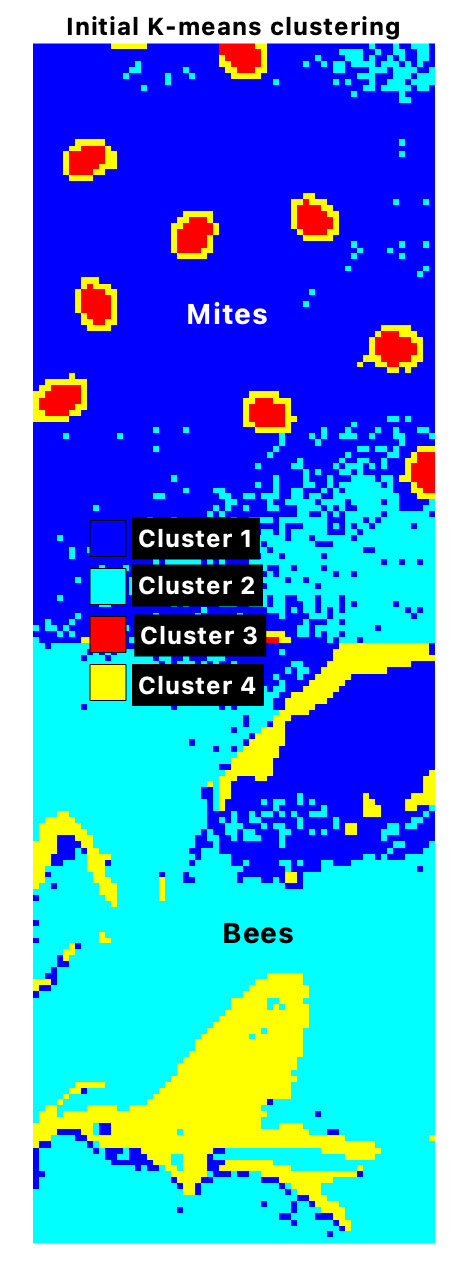}
        \caption{}
        \label{fig:calibrationResults}
    \end{subfigure}
    \begin{subfigure}[b]{0.59\linewidth}
        \centering
        \includegraphics[width=\linewidth]{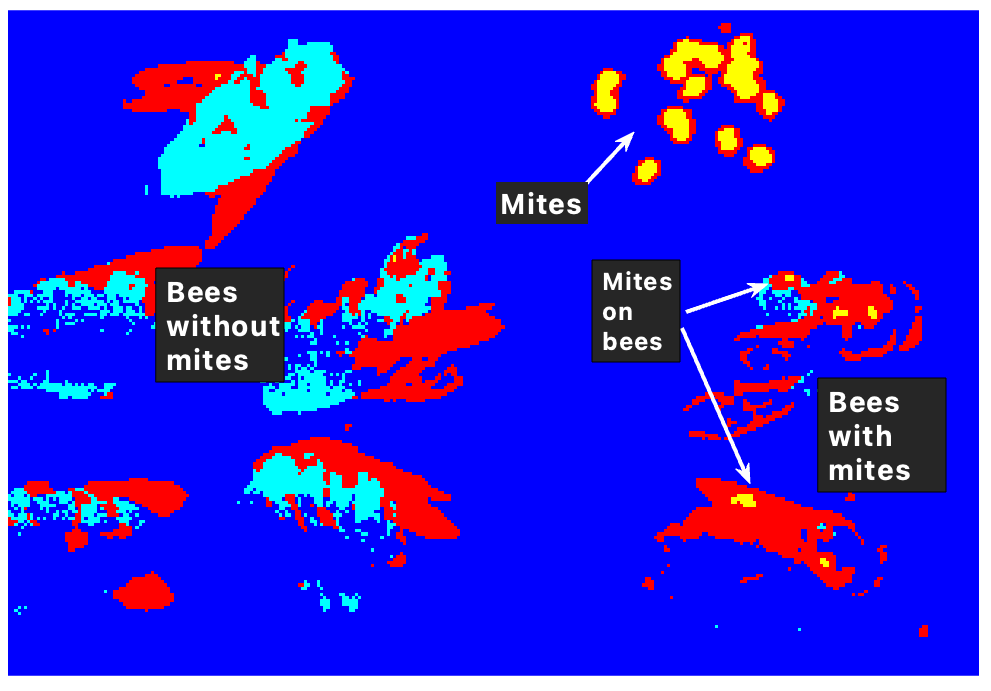}
        \caption{}
        \label{fig:testingResults}
    \end{subfigure}
    \caption{(a) Calibration and (b) testing image k-means clustering results in four clusters on full reconstructed spectra (Note: Cluster colors are not consistent between plots, red and yellow have been switched for visibility in the right-hand plot).}
    \label{fig:fullWavelengthResults}
\end{figure}

A minimum of four clusters are necessary for the k-means algorithm to separate the Varroa mites from the bees, as seen in Fig.~\ref{fig:calibrationResults}. If the clustering is ran with a lower number of clusters, the Varroa mite and some parts of the bee are being clustered together. 

The full-wavelength model utilizes the spectral reconstruction of the second and third principal components as an input to the k-means algorithm. The testing image (Fig.~\ref{fig:testingResults}) has been centered and scaled with the calibration image mean and standard deviation prior to the PCA projection and spectral reconstruction.

Furthermore, Fig.~\ref{fig:fullWavelengthResults} shows that with four clusters, the bees are not fully separated from the background, yet the Varroa mites are already separated into their cluster. The mites can even be detected on top of the bees Fig.~\ref{fig:testingResults}.


\begin{figure}[H]
    \centering
    \begin{subfigure}[b]{0.49\linewidth}
        \centering
        \includegraphics[width=\linewidth]{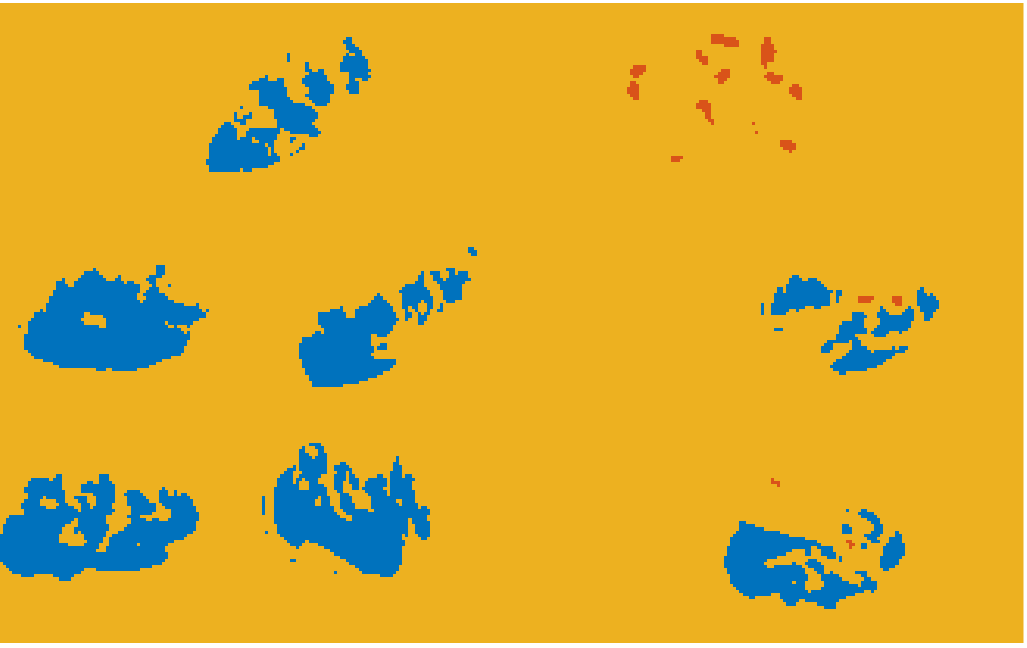}
        \caption{}
        \label{fig:KFPLSResults1}
    \end{subfigure}
    \hfill
    \begin{subfigure}[b]{0.49\linewidth}
        \centering
        \includegraphics[width=\linewidth]{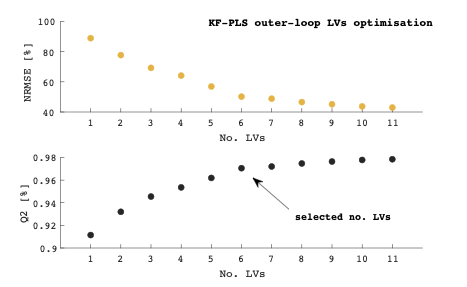}
        \caption{}
        \label{fig:KFPLSLVs}
    \end{subfigure}
    \caption{The KF-PLS results for a Matern5/2 Kernel function on the test image (a). The number of optimal LVs in the KF-PLS was 6. After the optimal number of LVs, the identification of Varroa mites was still successful.}
    \label{fig:KFPLSResults}
\end{figure}

The KF-PLS algorithm needed 150 iterations to converge. It was trained on four classes containing 300 pixels from each category: Varroa mite, bee wing, bee body and background. The results of the KF-PLS on the test image are seen in Fig.~\ref{fig:KFPLSResults}. The hyperparameters utilized to obtain the results were a learning rate of 0.1, Polyak's momentum for parameter update, 20 sub-samplings per iteration, and a batch sampling ratio of 50\% of the observations for each iteration. The Kernel function utilized was Matern5/2. As shown in Fig.~\ref{fig:KFPLSLVs} the KF-PLS needed 6 latent variables to converge, which was confirmed in the external loop evaluated with the optimized learned kernel parameters. 

\subsection{Wavelength selection}

Due to the inconsistent variation of the last 10 wavelengths, they have not been included in the list of selected variables, regardless of the wavelength selection method. The variable's value varied more with the spatial location of the pixel and not the type of material measured.

\begin{figure}[H]
    \centering
    \begin{subfigure}[b]{0.50\linewidth}
        \includegraphics[width=\linewidth]{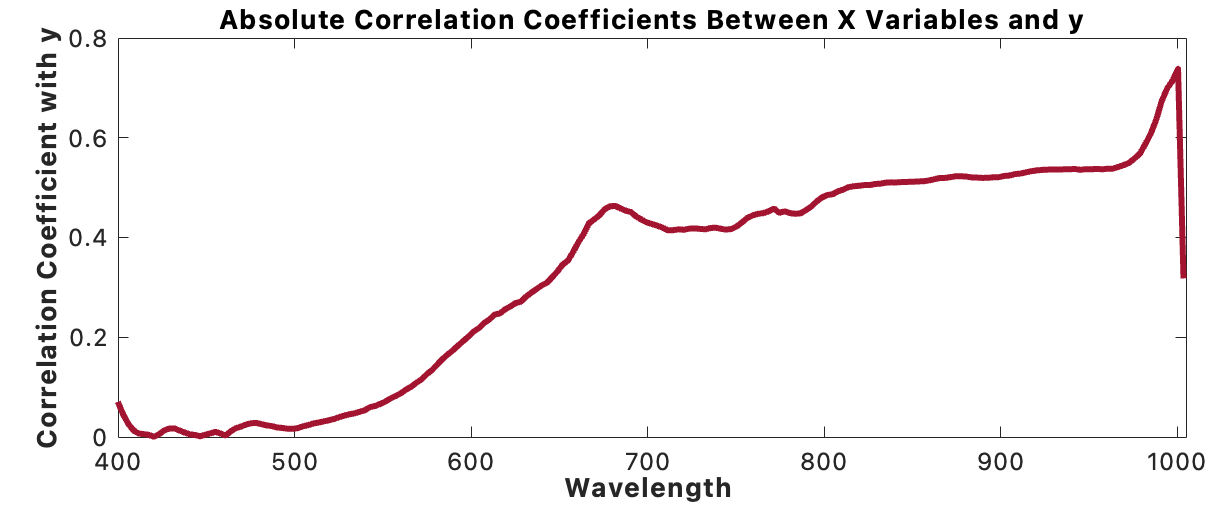}
        \caption{}
        \label{fig:correlationC}
    \end{subfigure}
    \hfill
    \begin{subfigure}[b]{0.47\linewidth}
        \includegraphics[width=\linewidth]{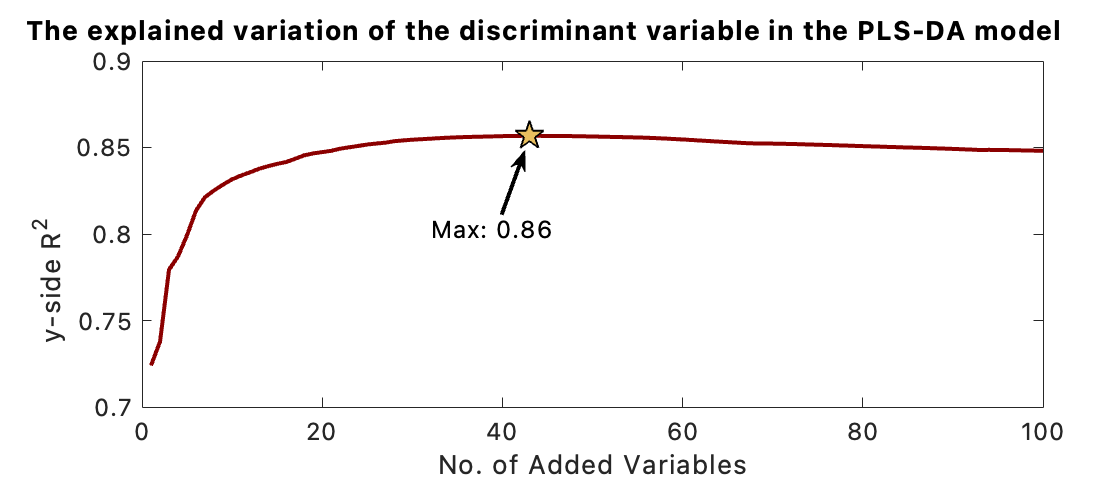}
        \caption{}
        \label{fig:PLSDAR2}
    \end{subfigure}

    \begin{subfigure}[b]{0.49\linewidth}
        \includegraphics[width=\linewidth]{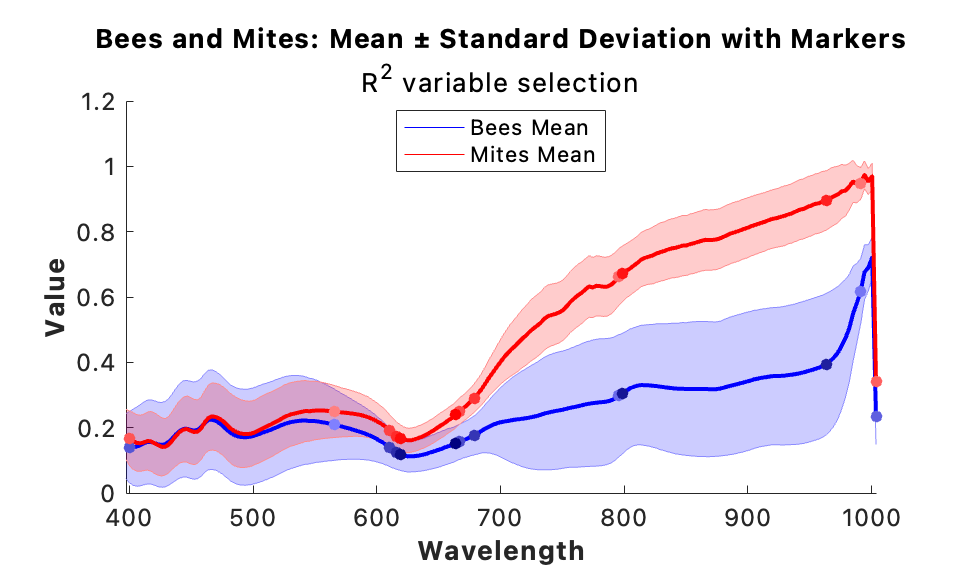}
        \caption{}
        \label{fig:R2varselect}
    \end{subfigure}
    \hfill
    \begin{subfigure}[b]{0.45\linewidth}
        \includegraphics[width=\linewidth]{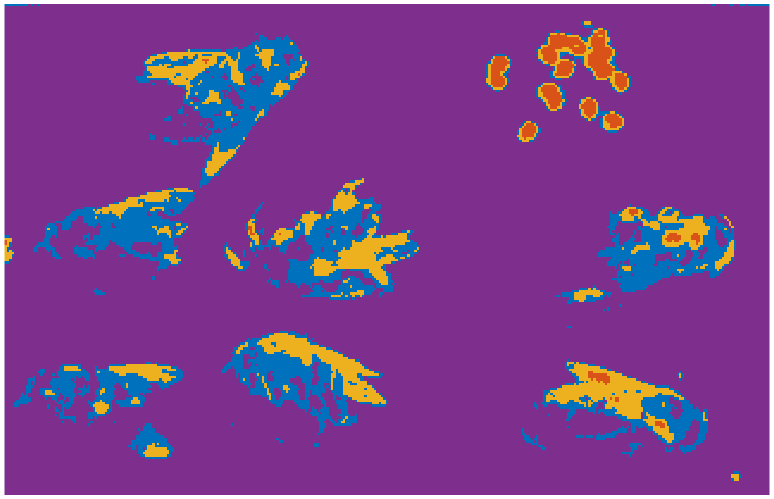}
        \caption{}
        \label{fig:R2result}
    \end{subfigure}
    
    \caption{(a) The absolute correlation coefficients between wavelengths and y-variable - the initializing step of the PLS-DA model, whose explained variance increases with the number of added variables is present in figure (b). (c) the minimum number of variables to correctly identify the independent data set (d).}
    \label{fig:R2wavelength}
\end{figure}

In the $R^2$ variable selection method, the initializing step requires analyzing the correlation coefficients of individual wavelengths with the response variable. The absolute value of the correlation coefficients is showcased in Fig.~\ref{fig:correlationC}. The higher the wavelength, the higher the absolute correlation with the discriminating variables. The three more-correlated variables were included in the initialization vector for the $R^2$-based selection. 

Even though the maximum $R^2$ value of 86\% was obtained in a PLS model with approx. 40 variables (Fig.~\ref{fig:PLSDAR2}, only the first 12 selected variables were necessary to obtain the test partition discrimination correctly. The 12 essential wavelengths can be observed in Fig.~\ref{fig:R2varselect}. A notable selection is the one at 800nm, which will be important in the further discussed results.

As observed in Fig.~\ref{fig:R2result}, Varroa mites could be identified, even if they were placed on top of the bees, but two false alarms were observed as well: one in the wing of the upmost bee and another one in the leg of another bee. These errors are presented as singular pixels and can be processed, if needed, with image processing techniques such as image erosion. 

\begin{figure}[H]
    \centering
    \begin{subfigure}[b]{0.7\linewidth}
        \includegraphics[width=\linewidth]{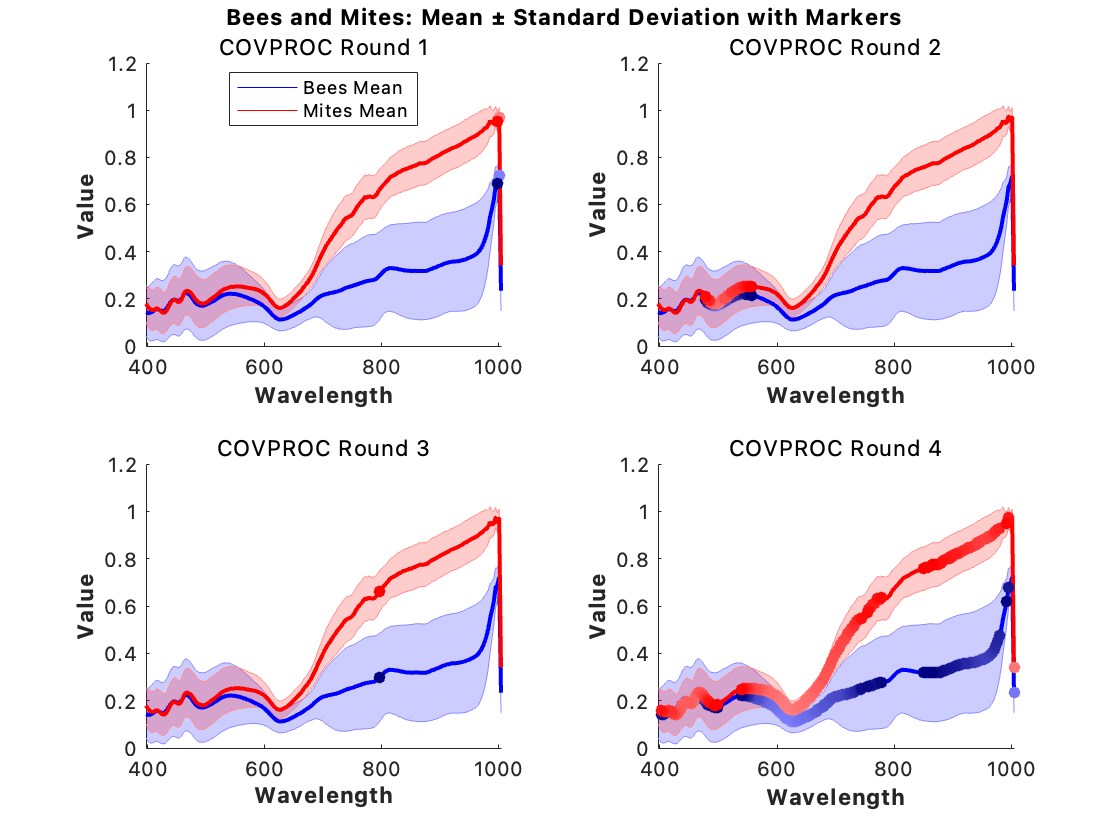}
        \caption{}
        \label{fig:COVPROCRounds}
    \end{subfigure}
    \begin{subfigure}[b]{0.5\linewidth}
        \includegraphics[width=\linewidth]{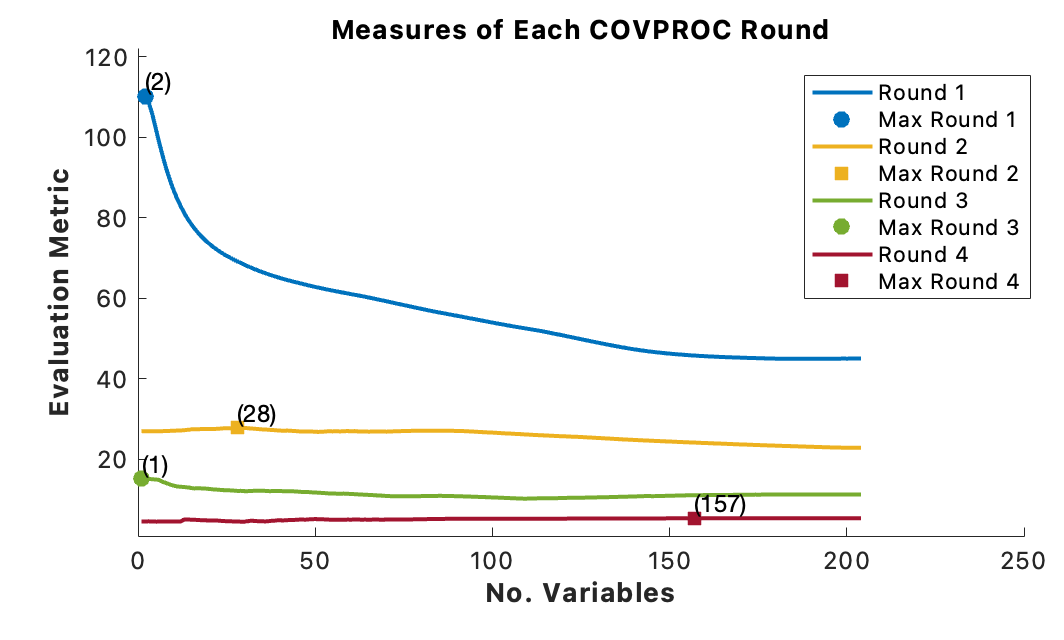}
        \caption{}
        \label{fig:COVPROCLines}
    \end{subfigure}
    \hfill
    \begin{subfigure}[b]{0.43\linewidth}
        \includegraphics[width=\linewidth]{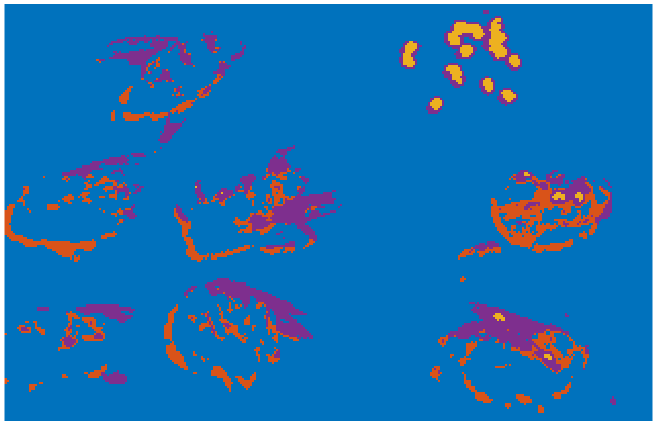}
        \caption{}
        \label{fig:COVPROCResults}
    \end{subfigure}
    \caption{(a) The variables selected in each COVPROC round. A light color in the marker indicates that the variable was selected early in the selection process. A darker marker indicates that the variable was selected later in the round. The evaluation metric results for each round are found in Fig. (b). }
    \label{fig:COVPROC}
\end{figure}

The second variable selection method compared, COVPROC, has the variable selection done in rounds. Fig.~\ref{fig:COVPROCLines} shows the evaluation metric $\alpha$ for each of the rounds and the number of variables necessary for the peak of the evaluation metric.

As seen in Fig.~\ref{fig:COVPROCRounds}, two variables that were in the exclusion area were selected in the first round. In the second round, a non-intuitive list of 28 variables from the beginning of the spectrum was selected. The spectral region selected presents the highest overlaps in the distributions of bees and Varroa mites. In round three, only one variable, around 800 nm, was responsible for reaching the maximal evaluation metric. In contrast, round 4 presents an exhaustive list of 157 variables, indicating that further evaluations should not be considered. 

With the 1st round excluded, the variables selected in the 2nd and the 3rd round were enough to yield a good classification. It was observed practically that adding the 3rd round before the 2nd gave even better results. Instead of the 29 necessary variables in the round 2+3 succession, only four variables were needed if the selection order was round 3+2. This resulted in the image Fig.~\ref{fig:COVPROCResults} being obtained with the spectral reconstruction of only four wavelengths, one around 800~nm and three consecutive wavelengths around 500~nm.

\subsection{Discussion on future research} %

This is the first time hyperspectral images have been utilized to detect Varroa mites on bees' bodies. Studies with multispectral cameras have been done in literature, such as Berjge et al., 2019~\cite{BJERGE2019104898}, where the camera had 19 spectral bands. In the current study, 204 bands were utilized, a tenfold increase. This provides more robustness to the analysis and provides more information on what wavelengths would best discriminate between bees and mites. These wavelengths were identified to be 492.97~nm, 498.8~nm, 507.56~nm and 796.74~nm. These are similar to bands 470~nm, 630~nm and 780~nm identified in \cite{BJERGE2019104898}. However, based on the HS data, the 630~nm band gives inconsistent results, which can lead to false alarms, confusing pollinated bee's legs as mites.

Precise identification of discriminating wavelengths is crucial for developing a real-time monitoring system in which bees enter the hive through a limited passway, as suggested in~\cite{BJERGE2019104898}, which is continuously captured by a camera. With suitable monochromatic illumination and a common camera with a custom filter, Varroa mite-infested bees could be easily distinguished without the need for an expensive multispectral camera. Such a device could be used for long-term and regular beehive monitoring, which could be more reliable in Varroa mite detection than broad-spectrum white illumination. Sensing in such illumination could be challenging due to high color similarity between the mite and bee or due to partial occlusion of the mite between the bee's abdomen segments. For this purpose, we plan to improve and test our device under development, partially described in~\cite{BUT184291} as a part of future research based on the findings of this study.



The current study considered only discrimination between bees and mites, as that would be the main application. Thus, the rest of the hive detritus has been left unstudied. The HS images could reveal more information from this fraction as well. Our proposed approach could be used for the detritus analysis, including Varroa mites detection as suggested in~\cite{9897809}. This would require further development of the method to distinguish between other objects in detritus, such as wax, not just the bees and Varroa mites. However, in the proposed use case, the bees entering the hive should be groomed and thus not carrying any wax. Nevertheless, further methodology development is planned to analyse the next in-field measurements.

\section{Conclusion}

The present article demonstrated that hyperspectral imagery can be utilized in beehive health monitoring and proposed mathematical procedures for achieving Varroa mite identification utilizing hyperspectral imagery. The research revealed that unsupervised (K-means) and supervised (KF-PLS) clustering methods are efficient in hyperspectral parasite identification. The unsupervised methods require more spectral pre-processing, such as the reprojection from the most correlated principal components with a discriminative variable. Without a multivariate profile segmentation, clear parasite/bee clusters are not obtained, and the number of clusters to place the insects and their parasites in different groups is heavily increased. This occurs due to the largest amount of information being utilized to discriminate between objects and background, and often the background is not static, thus unreliable. The unsupervised method (KF-PLS) proved efficient for bee-mite discrimination without significant preprocessing, and the best convergence time was obtained using a Matern 5/2 kernel function. 


Two methods for finding the wavelengths that best discriminate between bees and mites are the coefficient of determination ($R^2$)- based method and the COVPROC method.  Both proved high wavelength reduction from the initial 204-wavelength set: the COVPROC method reduced the wavelengths to 4 essential ones, whereas the $R^2$ method resulted in 12 wavelengths selected that were able to give a good discrimination in the unsupervised methodology. Even if the wavelength selection was done, re-projection from spectral profiles still needed to achieve good clustering. No raw spectra with a certain number of selected bands could correctly distinguish between all Varroa mites and bees, except for the supervised KF-PLS method.


The current HS dataset has been made publicly available to support further research in this field. In future research, the dataset will be expanded with more in-field measurements. The methodology will also be expanded to distinguish between more classes of bee detritus (not only bees and Varroa mites but also wax, pollen, sugar, and other debris). Furthermore, a deeper analysis of the methods' discrimination ability on a more complex scene is needed. The results of discriminating wavelengths can already be applied to the real-time monitoring setup. 


\section*{Acknowledgements}
Funding from Research Council of Finland for Centre of Excellence of Inverse Modelling and Imaging, project number 353095, is acknowledged, and the Research Council of Finland through the Flagship of Advanced Mathematics for Sensing, Imaging and Modelling (decision number 359183) is also acknowledged. The presented project was further supported by the grant no. FEKT-S-23-8451 "Research on advanced methods and technologies in cybernetics, robotics, artificial intelligence, automation and measurement" from the Internal science fund of Brno University of Technology. 


\end{document}